\documentclass{article}
\usepackage[final]{neurips_2022}

\usepackage[utf8]{inputenc} %
\usepackage[T1]{fontenc}    %
\usepackage{hyperref}       %
\usepackage{url}            %
\usepackage{booktabs}       %
\usepackage{amsfonts}       %
\usepackage{nicefrac}       %
\usepackage{microtype}      %
\usepackage{xcolor}         %
\usepackage{graphicx}
\usepackage{enumitem}
\setcitestyle{square,numbers}

\graphicspath{ {./images/} }

\usepackage{booktabs}
\newcommand{\ra}[1]{\renewcommand{\arraystretch}{#1}}

\usepackage{amssymb}
\usepackage{pifont}
\newcommand{\xmark}{\ding{55}}%

\newcommand\blfootnote[1]{%
  \begin{NoHyper}%
  \begingroup
  \renewcommand\thefootnote{}\footnote{#1}%
  \addtocounter{footnote}{-1}%
  \endgroup
  \end{NoHyper}%
}
\begin{document}
\title{ On the Surprising Effectiveness of Transformers in Low-Labeled Video Recognition}

\author{%
Farrukh Rahman$^{12*}$ \quad Ömer Mubarek$^{13*}$ \quad Zsolt Kira$^1$\\
$^1$Georgia Institute Of Technology \quad $^2$Microsoft \quad $^3$HERE Technologies\\
}

\maketitle
\thispagestyle{empty}

\begin{abstract}
  \looseness=-1 Recently vision transformers have been shown to be competitive with convolution-based methods (CNNs) broadly across multiple vision tasks. The less restrictive inductive bias of transformers endows greater representational capacity in comparison with CNNs. However, in the image classification setting this flexibility comes with a trade-off with respect to sample efficiency, where transformers require ImageNet-scale training. This notion has carried over to video where transformers have not yet been explored for video classification in the low-labeled or semi-supervised settings. Our work empirically explores the low data regime for video classification and discovers that, surprisingly, transformers perform extremely well in the low-labeled video setting compared to CNNs. We specifically evaluate video vision transformers across two contrasting video datasets (Kinetics-400 and SomethingSomething-V2) and perform thorough analysis and  ablation studies to explain this observation using the predominant features of video transformer architectures. We even show that using just the labeled data,  transformers significantly outperform complex semi-supervised CNN methods that leverage large-scale unlabeled data as well. Our experiments inform our recommendation that semi-supervised learning video work should consider the use of video transformers in the future. 
\end{abstract}
{\blfootnote{\footnotesize \textsuperscript{*}Equal Contribution}}

\section{Introduction}

 \looseness=-1 Deep learning has achieved significant success in various video recognition tasks.  Recent improvements have been driven by switching from convolutional neural networks (CNNs) to transformer architectures. The success of transformers on tasks in natural language \cite{devlin2018bert} and machine translation \cite{vaswani2017attention} has inspired their adaptation for visual tasks such as image classification \cite{dosovitskiy2020ViT, liu2021swin} and object detection \cite{carion2020end}. The attention mechanism \cite{bahdanau2014neural} used by transformers is effective at learning long range relations between sets of inputs which makes them a natural extension to spatio-tempororal video tasks as video is also sequential. Recent work has realized this potential and Video Transformers (VTs) have been effective on a range of video tasks, including activity recognition. %

 \looseness=-1 Just like for image classification, the application of transformers to videos has largely been driven by large-scale datasets ~\cite{kay2017kinetics, goyal2017something}. Indeed, transformers are more flexible models with a less restrictive inductive bias, leading to greater representational capacity compared to CNNs. This leads to data-efficiency concerns~\cite{khan2021transformers}. While specialized architectures can somewhat ameliorate this~\cite{liu2021efficient,el2021large}, the dominant usage still involves ImageNet-scale training for image classification (at least ImageNet-1K or IN1k) and Kinetics-scale training for videos. 
 
 \begin{figure}
 \resizebox{\linewidth}{!}{
  \includegraphics[width=\linewidth]{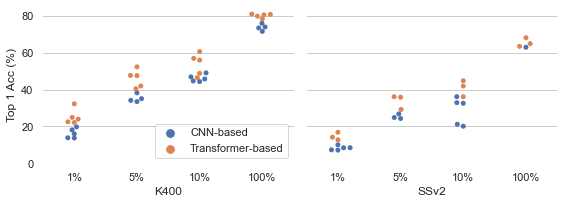}}
  \caption{Transformers-like vs CNN-like architectures in low-labeled video classification settings for K400 and SSv2. While the gap at 100\% is small, the gap increases as labeled training data decreases across both datasets. Note the higher variance of Transformer based architectures compared with CNNs. This is likely due to the variance in architecture designs between CNNs and Transformers commonly used for video recognition.}
  \label{fig:figA1VtvsCNN}
\end{figure}

 \looseness=-1 It is therefore not clear whether transformers would be suited to the challenging low-labeled video setting. In our work, we perform a thorough empirical analysis on the use of vision transformers for video classification in the low labeled regime. 
Through experiments, we arrive at the finding that despite the popularity of CNNs in such works, and the intuition that transformers often require more data, \textit{transformer-based models far surpass convolution-based architectures in low-labeled supervised video classification} across multiple datasets (Figure~\ref{fig:figA1VtvsCNN} \& Tabs.~\ref{tab:K400COMP}-~\ref{tab:SSv2COMP}) 

\looseness=-1 Based on this finding, we conduct a rigorous investigation into which types of transformers are especially suited to this task, and why.
Specifically, we study Video Swin Transformer \cite{liu2021videoswin}, Uniformer \cite{li2022uniformer3d}, ViViT \cite{arnab2021vivit}, and MViT \cite{li2021improvedMViT} over K400 \cite{kay2017kinetics} and SSv2 \cite{goyal2017something} in the low-labeled setting alongside several CNN based architectures.
Our observations lead to ablations for Video Swin for the key components such as shifted window, relative position bias, and spatial pretraining on 1N1k \cite{deng2009imagenet}. 
Through this we identify spatial pretraining on IN1k as the primary source for the observed performance gap with CNNs. 
We further modified Video Swin transformer to explicitly mimic Uniformer's Local-Local-Global-Global attention pattern which leads to slight improvements on K400.

Further, one setting that utilizes less labels along-side larger-scale unlabeled data is semi-supervised learning. 
 \looseness=-1 Among existing semi-supervised video classification works~\cite{sign2021tcl, xiong2021multiview, xiao2022temporalgradlearning, terao2022compressedssl, jing2021videossl}, the focus has tended to be on architectures based on convolutional neural networks (CNNs), owing to the rapidly expanding set of video architectures of this type. While transformers have set new fully-supervised benchmarks, surprisingly, \textit{none of the recent semi-supervised video classification works have explored transformer-based architectures for this problem}. %
 One reason is that transformers are thought to be data-inefficient based on image classification works~\cite{liu2021efficient, dosovitskiy2020ViT, khan2021transformers}. Moreover the attention operation's quadratic cost is compounded by the temporal dimension in video. Recent work ameliorates this limitation with VTs by re-injecting a local inductive bias \cite{liu2021swin, li2022uniformer2d, li2022uniformer3d}, factorizing the space-time attention mechanism \cite{dosovitskiy2020ViT, bertasius2021space}, and/or using hierarchical feature learning \cite{liu2021swin, fan2021multiscale}. We show that transformers using \textit{labeled data alone} can beat all existing complex semi-supervised CNN-based methods, which leverage the larger-scale unlabeled as well. This result suggests that the community should focus on such architectures, in addition to other orthogonal algorithmic improvements as transformer-based models can serve as a strong foundation for developing semi-supervised methods for video classification tasks.  

In summary, we make the following novel observations:

\begin{enumerate}
    \item Video Swin Transformers outperform CNNs and other video transformers overall for low-labeled video classification.
    \item Through ablations we discover initialization with image level pretraining is the key factor for the success of vision transformers in video.
    \item  Modifying VST to be more global in the latter stages improves its low-labeled performance marginally on K400 while remaining competitive on SSv2, however cannot make up the gap with Uniformer.
    \item Despite their popularity in semi-supervised video classification, CNNs are significantly outperformed by Vision transformers over multiple types of video datasets, without even leveraging the unlabeled data.
\end{enumerate}

\section{Related Work}
\label{related}

\subsection{CNNs for Video recognition}
\looseness=-1  The success of CNNs on images led to natural early extensions on video \cite{taylor2010convolutional, kim2007human, jhuang2007biologically}. Video is a sequential task where spatial information changes temporally. Video frames are temporally correlated leading to information redundancy. One approach is to reduce the problem to a single frame problem for feature extraction and apply pooling on the sequence of features \cite{karpathy2014largeVideo}. This pooling, however, loses temporal information. To learn finer spatio-temporal features, prior works \cite{ji2012-3d, baccouche2011sequential3dcnnlstm, tran2015c3d, karpathy2014largeVideo} extend 2D spatial convolution to 3D spatiotemporal convolutions. 

Simonyan et al. \cite{simonyan2014twostream} use both RGB with optical flow through two stream networks improves both performance and efficiency and Feichtenhofer et al. \cite{christoph2016spatiotemporal, feichtenhofer2017spatiotemporal} extend this approach by adding residual connections \cite{he2016resnet} in the architecture or fusion between the streams. Tran et al. \cite{tran2018r3d, tran2019video} observe 3D spatio-temporal convolutions can be factorized to 2D spatial followed by 1D temporal convolution (R(2+1)3D) increasing expressivity of the model (interleaving more nonlinear activations) while benefiting optimization. Carreira et al \cite{carreira2017quoi3d} improve the efficiency of 3D convolutions by using inception \cite{szegedy2015going} and bootstrapping 3D filters from 2D CNNs trained on Imagenet \cite{deng2009imagenet}. TSM (temporal shift module) \cite{lin2019tsm} tackles efficiency by shifting temporal information allowing efficient mixing of information which can be used for both 2D and 3D CNNs.  
The notion that spatial characteristics in activity recognition change at a slower rate than motion motivated a two-stream method Slowfast \cite{feichtenhofer2019slowfast}. Slowfast incorporates two RGB streams, the first a slow pathway to capture the former and a faster pathway to capture the latter, and fuses them. Feichtenhofer et al. \cite{feichtenhofer2020x3d} design a method that starts from an efficient 2D architecture and expands itself across multiple axis. 

\subsection{Self-attention for Video Recognition}
\looseness=-1 Sequence modeling has often been paired with CNNs owing to the temporal nature of video; early works \cite{baccouche2011sequential3dcnnlstm, yue2015deepcnnlstm, donahue2015long} apply a sequence learner (eg. LSTM) over convolutions to learn the evolution of spatial features over time. The success of self-attention (SA) \cite{bahdanau2014neural, weston2014memory, graves2014neural} across various NLP tasks \cite{vaswani2017attention, devlin2018bert} has inspired its use in computer vision with success across multiple visual tasks \cite{wang2018non, guo2022attention}. SA acts as a method to dynamically weight features based on the input, conferring the capacity to flexibly select aspects of the input that are more relevant to the task. This naturally extends to video analysis where spatial, channel-wise, or temporal attention may be relevant breaking past the local inductive bias imposed by convolution. SA has been applied on video alongside a sequential learner such as a LSTM \cite{li2018videolstm} or directly with CNN based feature maps \cite{wang2018non, sharma2015action, long2018attention, chen20182}. Recent success with SA as a standalone layer for vision \cite{dosovitskiy2020ViT, touvron2021training} has been extended to video \cite{arnab2021vivit, li2022uniformer3d, liu2021videoswin, bertasius2021space, bulat2021space, fan2021multiscale, neimark2021video, li2021improvedMViT}. SA is an operation that scales with input size making computation with high dimensional spatiotemporal inputs inefficient. An approach introduced by ViT \cite{dosovitskiy2020ViT} image domain is to patchify and linearly embed the input which has also been adopted for other video transformer architectures. While this ameliorates SA's complexity by reducing the number of visual tokens, other approaches have been explored to further reduce complexity. These include factorizing spatial and temporal attention allowing for specialization along either of the axes \cite{arnab2021vivit, bertasius2021space, bulat2021space}, hierarchical representation with pooled attention \cite{fan2021multiscale, li2021improvedMViT} or adding convolution-like inductive bias \cite{liu2021videoswin, li2022uniformer3d}.

\subsection{Semi-supervised learning for low labeled video classification}
\looseness=-1 The semi-supervised learning (SSL) setting considers the scenario where copious unlabeled examples are present with some smaller amount of labeled data (1\%, 5\%, or 10\% are typical labeled splits). A naive approach might be to simply use the labeled data only, discarding the unlabeled data. However, SSL seeks to improve upon this by using both labeled and unlabeled data. The success of SSL in image classification \cite{chen2020big, sohn2020fixmatch} has been recently extended to video classification \cite{sign2021tcl, xiong2021multiview, xiao2022temporalgradlearning, terao2022compressedssl, jing2021videossl, kumar2022end, rizve2021defense}. These methods, however, use convolution-based backbones; self-attention based methods such as video transformers have not been explored to our knowledge.

\section{Preliminaries - Vision Transformers}
\label{prelim}
\looseness=-1 Given their success in NLP and machine translation tasks, recently transformer based architectures have been applied to image and video tasks. A fundamental vision transformer architecture is ViT \cite{dosovitskiy2020ViT} which is a near direct transfer of \cite{vaswani2017attention} for image classification.  Specifically, instead of attending over image pixels, ViT divides an image into non-overlapping patches, applies a linear projection to these image patches and converts these 2D patches into a sequence of 1D tokens. This sequence of linear embeddings is then passed to the transformer. The success of ViT has led to numerous transformer inspired architectures for video classification. ViViT \cite{arnab2021vivit}, a straightforward extension of ViT to video, divides and projects 3D patches to a sequence of tokens. Global self-attention has a quadratic $O(n^2)$ computational cost with respect to the input sequence length $n$. This bottlenecks the size of the context window allowable for sequence modeling. For text, this bottleneck can be mitigated partly by tokenizing text through sub-word tokenization \cite{sennrich2015subword} or byte pair encoding \cite{gage1994BPE}, ie. compressing the sequence into a smaller set of tokens. Self-attention over pixels is intractable,  hence images are broken up into non-overlapping patches and embedded as visual tokens; a procedure which naturally extends to video at the cost of additional complexity due to increased dimensionality of each spatio-temporal token. Recent approaches for vision transformers seek to use stronger inductive biases such as local windows and hierarchical representation to further reduce computational complexity while retaining modeling flexibility. 

\looseness=-1 We focus on empirically studying various VTs architectures: Video Swin Transformer \cite{liu2021videoswin}, Uniformer \cite{li2022uniformer3d}, ViViT \cite{arnab2021vivit}, and MViT \cite{li2021improvedMViT} are chosen to capture different architectural styles that have shown  high performance on the Kinetics-400 \cite{kay2017kinetics} and Something-Something v2 \cite{goyal2017something}. ViViT extends ViT to video using global self-attention between all spatio-temporal tokens. Uniformer and Video Swin Transformer (VST) on the other hand introduce inductive biases, namely replacing or combining global with local self-attention. Uniformer combines the two by using local attention in a CNN-like manner for early stages followed by global self-attention during the latter stages. Video Swin Transformer completely replaces global attention with local and uses window shifting to mix tokens through local attention. %

\subsection{Vision Transformer Architectures}
\begin{figure*}
  \resizebox{\linewidth}{!}{
  \includegraphics[height=5cm, width=\linewidth]{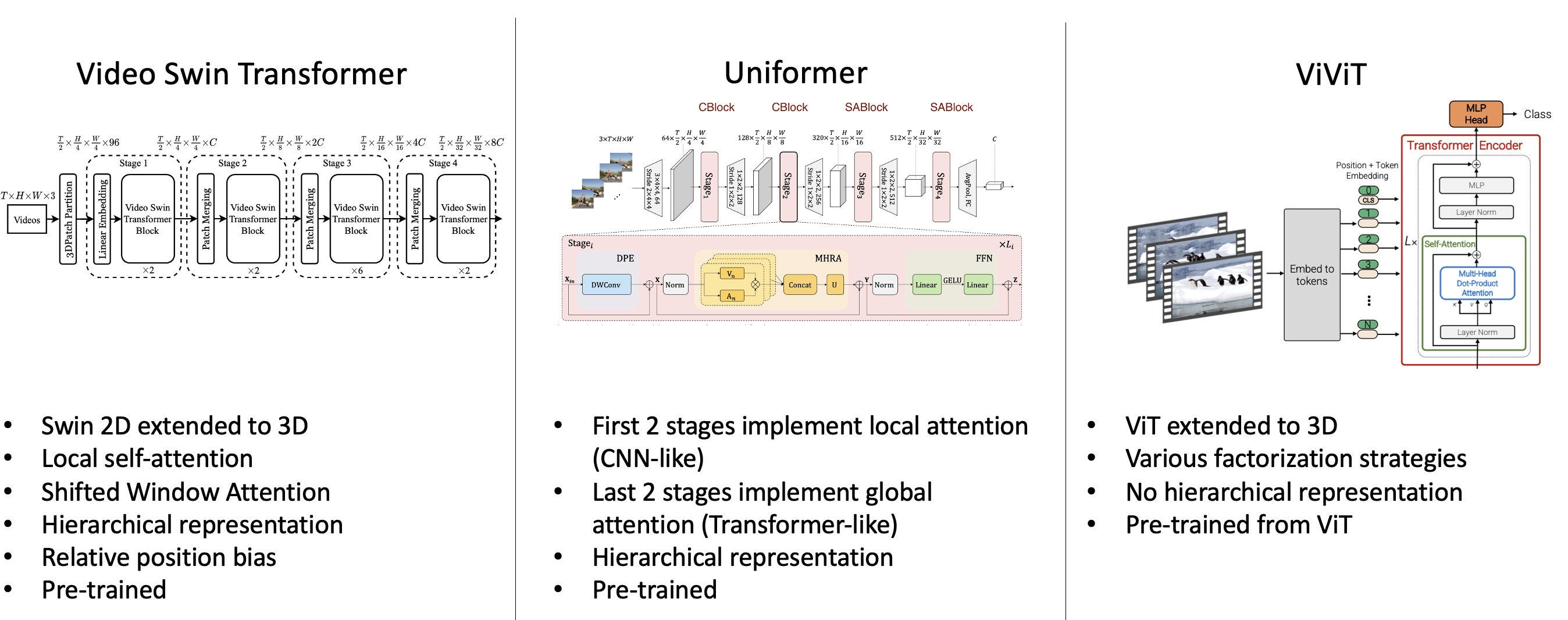}}
  \caption{Side-by-side comparison of main differences amongst the VT architectures: Video Swin Transformer \cite{liu2021videoswin}, Uniformer \cite{li2022uniformer3d} and ViViT \cite{arnab2021vivit}. Architecture diagrams taken from respective works.}
  \label{VtArchs}
\end{figure*}

\looseness=-1 Each vision transformer architecture tackles maintaining modeling flexibility with efficiency through related but different mechanisms. (See Fig.~\ref{VtArchs} for illustration of architectural comparison.)

\subsubsection{ViViT}
\label{sec:vivit}
\looseness=-1 ViViT \cite{arnab2021vivit} is an extension of ViT to video which computes global self-attention over spatio-temporal tokens. These tokens are generated by dividing the input to non-overlapping patches and are linearly embedded. Spatio-temporal inputs have greater number of tokens making global-self attention inefficient; as a result Arnab et al. studied several variants for increasing efficiency. We study two of these variants which were shown to have the highest performance on video classification tasks, ViViT Spatio-temporal (ST) and ViViT Factorized Encoder (FE).

\begin{itemize}[leftmargin=*]
\item \textbf{ViViT SpatioTemporal (ST):}
\label{sec:vivitST}
This is the default configuration, where global attention is computed over all of the spatiatemporal tokens, as described above.

\item \looseness=-1  \textbf{ViViT Factorised Encoder (FE):}
\looseness=-1 Factorized Encoder, in contrast to FSE, uses two encoders: One for spatial tokens which is effectively a ViT, and another for temporal. Global self-attention is computed for tokens \textit{within} an individual frame to obtain a spatial representation, which are computed for each frame in the video and  concatenated together to apply temporal self-attention.
\end{itemize}

\subsubsection{Video Swin Transformer}
Video Swin Transformer \cite{liu2021videoswin} (VST) adds several features to compensate for the lack of inductive bias in ViViT. For our experiments we used Swin Small for K400 and Swin Base for SSv2 (following \cite{liu2021videoswin}).

\begin{itemize}[leftmargin=*]
    \item \textbf{Local Self Attention:}
    Video Swin Transformer computes self-attention within non-overlapping local windows throughout its architecture. This limits the number of visual tokens a given query token attends to, increasing efficiency compared to traditional extensions like ViViT.
    
    \item \textbf{Shifted Window Attention:}
    The local windows in Video Swin Transformer do not have connections among each other, ie. tokens in one local window cannot attend to those in another. Therefore, while localized self-attention is more efficient, there is a tradeoff with reduced modeling power. To maintain the model’s capacity, Swin architecture implements shifted window partitioning in successive blocks.  Tokens from one local window are "shifted" or substituted with those from another window for successive local attention computation. This encourages mixing of information while maintaining the efficiency of local windowed attention, a concept similar to temporal shift module for CNNs \cite{lin2019tsm}.
    
    \item \textbf{Patch Merging Layers:}
    Swin Transformer creates CNN-like hierarchical representation via Patch Merging layers. As the architecture gets deeper, these layers reduce the number of tokens. The input token size $(T \times H \times W \times C)$ reduces to $(T \times H/2 \times W/2 \times 2C)$ all the way to $(T \times H/8 \times W/8 \times 8C)$. This effectively increases the reach of local attention throughout the network as the receptive field for a given token increases deeper into the network.
    
    \item \textbf{Relative Position Bias:}
    Instead of using absolute position embedding, Swin Transformer adds a relative position bias to each head during attention computation. This bias provides translation invariance like CNNs \cite{liu2021swin}.

\end{itemize}

\subsubsection{Uniformer}
At the heart of the Uniformer \cite{li2022uniformer2d} architecture lies Multi-Head Relation Aggregator (MHRA). Depending on how MHRA is instatiated lends itself to behave like a convolution-like operation or global self attention. This concept is naturally extended for spatio-temporal inputs \cite{li2022uniformer3d} by increasing the number of tokens similar to other architectures. Li et al. experiment with various configurations, eg. Local-Local-Local-Local vs Local-Local-Global-Global which denotes the instatiation for MHRA used. Local-Local-Global-Global (LLGG) was the best performing Uniformer configuration (performing 3D convolution over the visual tokens during early stages followed by global self attention in later stages). For both of K400 and SSv2, we used Uniformer Small configuration as it obtained very competitive performance and  similar results to the other transformers with 100\% labeled data.

\begin{itemize}[leftmargin=*]
\item \textbf{Shallow Layers:}
In the shallow layers/stages, where the number of tokens is high, Uniformer uses Local MHRA; learning from local spatiotemporal neighborhoods. Local MHRA is akin to 3D CNNs \cite{tran2015c3d, carreira2017i3d} 

\item \textbf{Deep Layers:}
\looseness=-1 When moving into deeper layers/stages, learning focuses on global content. This Global MHRA module resembles global self-attention in transformers. This is responsible for aiding in temporal modeling of longer-term relations between tokens. Overall Uniformer uses a similar notion of using CNNs for extraction of finer spatiotemporal (or 2D spatial) features followed by a form of aggregation, relational weighting, or further sequence modeling of said features \cite{wang2018non, li2018videolstm, baccouche2011sequential3dcnnlstm}.

\item \textbf{Hierarchical Representation:}
\looseness=-1 Similar to Swin Transformer \cite{liu2021swin}, Uniformer produces hierarchical representation by reducing the number of tokens as the network gets deeper.

\end{itemize}

\subsubsection{MViTv2: Multiscale Vision Transformer }
\looseness=-1 Improved Multiscale Vision Transformer (MViTv2) is a unified architecture for image and video classification \cite{li2021improvedMViT} that improves upon MViTv1 \cite{fan2021multiscale}. MViTv1 similar to transformers such as Uniformer \cite{li2022uniformer3d} and Swin \cite{liu2021videoswin,liu2021swin} in that they attempt to ameliorate training efficiency by adding inductive bias to ViT while maintaining model flexibility. MViT is motivated by the notion of multiscale feature hierarchy \cite{fan2021multiscale}; achieved by allowing the transformer to expand its channel dimensions while pooling spatiotemporal resolution.
\begin{itemize}[leftmargin=*]{}
\item \textbf{Multi Head Pooling Attention MHPA}
Multi-head Pooling attention endows the self-attention operation with a pooling operation over the Keys, Queries, Values ($K,Q,V$).  This pooling along each dimension reducing both the spatial and temporal resolution of the token representations which is fed into self-attention, improving efficiency of the self-attention operation \cite{li2021improvedMViT}.

\item \textbf{Decomposed Relative Position Embedding}
MViTv1 uses absolute positional embeddings to give self-attention some notion of positional relational information between visual tokens. This is common as the softmax in self-attention is permutation equivariant akin to selecting from a set. Absolute position affects relations between visual tokens to be dependent on where they occur in the input despite their relative positions being equal \cite{li2021improvedMViT}. MViTv2 uses relative position embedding to mitigate this by performing a pairwise spatiotemporal relative position embedding within the self-attention module.
\end{itemize}

\subsection{Datasets}
\label{datasets}
\looseness=-1 Our experiments were conducted on two widely-used large scale video action datasets: Kinetics-400 \cite{kay2017kinetics} and Something-Something v2 \cite{goyal2017something}. K400 consists of over 200K training examples across 400 classes. Each sample is a 10s long YouTube video clip that is human action centric,  i.e. focuses on human-human or human-object interactions. SSv2 has 220K videos, each 2-6 seconds duration over 174 classes. SSv2 is also human-object focused but is centered around everyday interactions between humans as objects; the object and background are more consistent in comparison with Kinetics. As such SSv2 requires more temporal modeling to distinguish finer motion, rather than leveraging frame-level static object information. This observation has been further quantified on mini-SSv2 and mini-Kinetics by Chen et al \cite{chen2021deep}. This contrast makes them good candidates for studying transformers vs CNNs in low-labeled scenarios.

\section{Experiments}
\label{questions}
\looseness=-1 We seek to understand empirically how VTs compare to CNNs under low labeled data settings for Action Recognition. Current intuition suggests that transformers, due to their flexibility and weaker inductive bias,  require large amounts of training data \cite{dosovitskiy2020ViT, khan2021transformers}. If true, it's important to measure the gap between the two as it will inform future research directions for both low-labeled and semi-supervised video understanding settings. CNN and transformer architectures are spatially pretrained on IN1K \cite{deng2009imagenet} for K400 and trained on IN1k and K400 100\% for SSv2.

\subsection{How do Vision Transformers compare to CNNs in low labeled video classification settings?}

 \looseness=-1 We compared several Video Vision Transformers (VTs) with CNNs in low labeled data settings on Kinetics-400 (K400) action recognition dataset using 1\%, 5\% and 10\% of training data, without using any unsupervised data. VST, Uniformer, MViTv2 and ViViT are compared against various purely Convolutional architectures with ResNet-50 \cite{he2016resnet} backbone trained on IN1K as baselines; namely R3D \cite{tran2018r3d}, I3D \cite{carreira2017i3d} which are both 3D CNNs as well as C2D~\cite{wang2018non}) which is a 2D CNN. I3D typically uses an additional Optical flow \cite{simonyan2014twostream} path in addition to RGB; this was not used which effectively made I3D a R3D. Moreover we pair these CNNs up with Non-local Networks \cite{wang2018non} conferring the architectures additional global expressivity similar to attention. 

 \looseness=-1 For K400 on 1\% split, VTs significantly outperform the CNNs; with VST \cite{liu2021videoswin} obtaining significant increase over the best CNN based model C2D-NL-R50 \cite{wang2018non} (32.3\% vs 19.79\%). Transformers perform better at 1\% over CNNs, eventually widening the gap before it closes close to 100\% labeled data.  As the percentage of labeled data increases from 1\% to 100\%, the gap between VTs and CNNs closes. This is a surprising observation as VTs are considered less sample efficient due to their more flexible modesling capacity with fewer inductive biases. However, our results indicate that spatially pretrained VTs outperform CNNs at low labeled settings.

\begin{table*}[ht]
    \ra{1.1}
    \begin{center}
        \scalebox{0.8}{
        \begin{tabular}{ccccccccc}
            \toprule
            \textbf{Architecture} &   \multicolumn{4}{c}{\textbf{Top-1 Accuracy}} & \textbf{\# Params (M)} & \textbf{Flops (G)} \\

             & 1\% & 5\% & 10\% & 100\%\\
            \toprule
            I3D-R50  \cite{carreira2017quoi3d}  & 13.73(\textcolor{red}{-3.79})  & 33.49 & 44.37(\textcolor{red}{-4.73})  & 73.5\textsuperscript{\textdaggerdbl} (\textcolor{red}{-2.5}) & 28.1 & 28.5\\
            I3D-NL-R50 \cite{wang2018non}  & 13.86 (\textcolor{red}{-2.14}) & 34.18 (\textcolor{red}{-0.69}) & 44.7 (\textcolor{red}{-4.4}) & 74\textsuperscript{\textdaggerdbl} (\textcolor{red}{-2}) & 35.4 & 36.7 \\
            C2D-R50  \cite{wang2018non} & 18.13 (\textcolor{red}{-1.66}) & 35.1 (\textcolor{red}{-3.13}) & 45.77 (\textcolor{red}{-3.33})  & 71.61\textsuperscript{\textdaggerdbl} (\textcolor{red}{-4.39}) & 24.3 & 19.6 \\
            C2D-NL-R50  \cite{wang2018non} & \textbf{19.79}  & \textbf{38.23}  & 46.93 ( (\textcolor{red}{-2.71}) & -  & 31.7 & 27.8\\
            R3D-50 from \cite{qian2021cvrl} & 16\textsuperscript{\textdagger} (\textcolor{red}{-2.13}) & - & \textbf{49.1}\textsuperscript{\textdagger}  & \textbf{76.0}\textsuperscript{\textdagger} & - & -  \\
            \hline
            ViViT FE \cite{arnab2021vivit} & 24.91 (\textcolor{blue}{+5.12}) & 41.98 (\textcolor{blue}{+3.75})& 48.84 (\textcolor{red}{-0.26}) & 78.4\textsuperscript{\textdagger} (\textcolor{blue}{+2.4}) & 115.1 & 284\\
            ViViT ST \cite{arnab2021vivit} & 23.99 (\textcolor{blue}{+4.2}) & 40.47 (\textcolor{blue}{2.24}) & 46.49 (\textcolor{red}{-2.61}) & 79.9\textsuperscript{\textdagger} (\textcolor{blue}{+3.9}) & 88.9 & 455.2\\
            Uniformer Small \cite{li2022uniformer3d} & 22.52 (\textcolor{blue}{+2.73}) & 47.71 (\textcolor{blue}{+9.48}) & 56.01 (\textcolor{blue}{+6.91}) & 80.8\textsuperscript{\textdagger} (\textcolor{blue}{+4.8}) & 21.4 & 167.2 \\
            MViTv2 Small \cite{li2022uniformer3d} & 22.17 (\textcolor{blue}{+2.38}) & 47.59  (\textcolor{blue}{+9.36}) & 56.96 (\textcolor{blue}{+7.86}) & 81.0 \textsuperscript{\textdagger} (\textcolor{blue}{+5}) & 34.5 & 64.5\\
            Video Swin Small \cite{liu2021videoswin} & 32.3 (\textcolor{blue}{+12.56}) & 52.37 (\textcolor{blue}{+14.14}) & 60.66 (\textcolor{blue}{+11.56}) & 80.6\textsuperscript{\textdagger} (\textcolor{blue}{+4.6}) & 49.8 & 166\\
        \bottomrule
        \end{tabular}
        }
        \caption{Top 1\% accuracy results on K400 for various low-labeled settings for CNN-based (top half) vs Transformer-based (bottom half). 
        Differences are computed against the highest performing CNN-based model for each split in \textbf{bold}. 
        Refer to supplemental for top-5 acc results. \textsuperscript{\textdagger}Result from respectively cited work. \textsuperscript{\textdaggerdbl}Result from \cite{fan2020pyslowfast}.}
                \label{tab:K400COMP}
    \end{center}
\end{table*}

 \looseness=-1 These results contradict the aforementioned intuition: not only is superior performance observed for VTs, but also they have a large gap at low labeled settings that increases with decreasing training data, Uniformer being the only exception. Our observations indicate methods that apply 3D convolution (Uniformer, I3D, R3D) are less data efficient (even when they reach similar performance at 100\%) in comparison with 2D CNNs or Swin Transformer. The latter, when spatially pretrained, transfers better to spatially dominant datasets like Kinetics. In contrast, on SSv2 3D convolutions pick up and Uniformer outperforms the others (Tab. \ref{tab:SSv2COMP}).  

\subsection{How do VTs compare among themselves?}
\label{VTselfcompare}
\looseness=-1 A comparison of VT performance in Tab.~\ref{tab:K400COMP} shows that in low-labeled settings Swin Transformer outperforms all other transformers by far on K400. The difference is highly pronounced at 1\% split: Swin at 32.3\%, the runner up (ViViT FE) at 24.91\%. While the Uniformer and the two variants of ViViT have very similar performance at 1\%, Uniformer dominates the ViViT variants at 5\% and 10\% splits. 

\looseness=-1 The superior performance of Swin over Uniformer is consistent with the superiority of VTs over CNNs: the first two of the four Uniformer stages behave like 3D convolutions. On the other hand, Swin is closer to a pure transformer-based architecture, which mixes its strong spatial pre-training with shifted attention. Both Swin and Uniformer %
outperform ViViT variants%
. The locality inductive biases Swin and Uniformer inject seem to help prevent noisy information from mixing from tokens far away. On the other hand, ViViT is susceptible to such noise as it computes global attention. 

\looseness=-1 A different picture arises in Tab.~\ref{tab:SSv2COMP}, which compares accuracies of VST and Uniformer for SSv2. At 1\% and 100\% of the training data, both VTs perform similar, with VST accuracy being slightly higher. However, for 5\% and 10\% splits, Uniformer clearly outperforms VST (\ref{tab:SSv2COMP}). This behaviour (compared to K400) can be attributed to the different nature of the datasets \cite{chen2021deep} . SSv2 challenges modeling of finer motions, as the background and objects are the same across multiple classes. 3D convolution is better able to capture finer spatio-temporal features as compared to sequence modeling. Uniformer seems to be able to leverage this as it use 3D convolutions during early layers resulting in better performance on SSv2.

\begin{table}[h!]

    \begin{center}
        
        \resizebox{0.7\linewidth}{!}{%
        \begin{tabular}{ccccccccc}
            \toprule
            \textbf{Architecture} & \multicolumn{4}{c}{\textbf{Top-1 Accuracy}} \\
             & 1\% & 5\% & 10\% & 100\% \\
            \hline
            I3D-R50  \cite{carreira2017quoi3d} & 8.47 (\textcolor{red}{-1.59})  & 24.3 (\textcolor{red}{-2.5}) & 32.92(\textcolor{red}{-3.25})  & -\\
            I3D-NL-R50  \cite{wang2018non} & 8.42(\textcolor{red}{-1.64})  & 24.85(\textcolor{red}{-1.95}) & 32.59(\textcolor{red}{-3.58}) & -\\
            C2D-R50  \cite{wang2018non} & 7.16 (\textcolor{red}{-2.9})  & - & 20.16 (\textcolor{red}{-16.01}) & -\\
            C2D-NL-R50  \cite{wang2018non} & 7.31 (\textcolor{red}{-2.75})  & - & 21.19 (\textcolor{red}{-16.04}) & -\\
            SlowFast-R50  \cite{feichtenhofer2019slowfast} & \textbf{10.06}  & \textbf{26.8} & \textbf{36.17} & \textbf{63}\textsuperscript{\textdaggerdbl}\\
            \hline
            Video Swin Base \cite{liu2021videoswin} & 14.3 (\textcolor{blue}{+4.11}) & 30.0 (\textcolor{blue}{+3.2}) & 37.5(\textcolor{blue}{+1.33}) & 65.67 (\textcolor{blue}{+2.67}) \\
            Uniformer Small \cite{li2022uniformer3d} & 12.77 (\textcolor{blue}{+2.9})  & 36.14 (\textcolor{blue}{+9.34}) & 44.81 (\textcolor{blue}{+8.64}) & 63.5\textsuperscript{\textdagger}(\textcolor{blue}{+0.5}) \\
            MViTv2 Small \cite{li2021improvedMViT} & 16.88 (\textcolor{blue}{+6.82})  & 35.88 (\textcolor{blue}{+9.08}) & 41.94 (\textcolor{blue}{+5.77}) & 68.2\textsuperscript{\textdagger}(\textcolor{blue}{+5.2}) \\
            \bottomrule
        \end{tabular}}
        \caption{Top 1\% accuracy on SSv2 for various low-labeled settings. CNN-based (top half) vs. Transformer-based (bottom half). Differences are computed against baseline  CNN-based model in \textbf{bold}. Refer to supplemental for full results. \textsuperscript{\textdagger}Result obtained from respectively cited work. \textsuperscript{\textdaggerdbl}Result obtained from \cite{fan2020pyslowfast}.} 
        \label{tab:SSv2COMP}

    \end{center}
\end{table}

\subsection{Why do VTs outperform CNNs at low labeled data settings?}
\looseness=-1 Our finding that VTs outperform CNNs in low-labeled video is nonintuive as it is commonly thought that transformers require large scale training to make up for their flexibility \cite{dosovitskiy2020ViT, liu2021efficient, khan2021transformers}.

\looseness=-1 VTs benefit from increased representation capacity due to self-attention layers: VTs use the attention layers to model global relations between tokens. On the other hand, receptive field of the CNNs locally limits the type of global relations which can be learned.

\looseness=-1 Swin transformer uses local attention via shifted local windows. These local windows are larger than CNN kernels and every second layer, the windows shift, such that attention is calculated between tokens not in the same original local window. This allows a token to compute self-attention with other tokens outside their local neighborhood. At the same time, local MSAs are known outperform global MSAs in VTs \cite{park2022howDoVTsWork}.

\looseness=-1 Uniformer employs local convolutions in the first two stages and switches to global MSA only after the resolution (i.e. number of tokens) is low enough. Both Swin and Uniformer inject some sort of locality inductive bias back into the transformer architecture.

\looseness=-1 Furthermore, these two architectures are multi-stage transformers, where the number of tokens reduces as the network gets deeper with each stage \cite{liu2021videoswin, li2022uniformer3d}. The result of this multi-stage architecture is hierarchical feature map resolutions like those in CNNs.
\subsection{Video Swin Transformer Ablation Study}
To better understand VST's performance and reasons described above, we conduct several ablations in low labeled data settings, see Tab.~\ref{tab:ablationsK400ssv2}. VST absent shifted windows is still competitive compared to CNNs across all splits and datasets. ie. spatial pretraining along with local window attention is still sufficient for VST to be better or competitive with CNNs across both K400 and SSv2.

\looseness=-1 VSTs implement local self-attention using shifted windows, employ relative position bias and are pretrained on ImageNet. In each of the ablations, we have stripped VST from only one of these features and examined the performance degradation from the full configuration. Ablation results for K400 and SSv2 are shown in Tab. \ref{tab:ablationsK400ssv2}.

Removal of any of the three features degrades the performance for all data splits, as shown in Tab.~\ref{tab:ablationsK400ssv2}.

\begin{table}[h!]
    \begin{center}
        \scalebox{0.9}{
        \begin{tabular}{ccccccc}
            \toprule
            \textbf{Ablation} & \multicolumn{3}{c}{\textbf{Top-1 Acc. K400}} & \multicolumn{3}{c}{\textbf{Top-1 Acc. SSv2}}\\
             & 1\% & 5\% & 10\% & 1\% & 5\% & 10\%\\
            \toprule
            None (Baseline) & 32.3 & 52.4 & 60.7 & 14.3 & 30.0 & 37.5\\
            \color{red}{\xmark} \color{black}\space Shifted Window & 28.6 (\textcolor{red}{-3.7}) & 50.3 (\textcolor{red}{-2.1}) & 57.7 (\textcolor{red}{-3.0})
            & 11.8 (\textcolor{red}{-2.5}) & 24.6 (\textcolor{red}{-5.4}) & 31.4 (\textcolor{red}{-6.1}) \\
            \color{red}{\xmark} \color{black}\space Relative Position Bias 
            & 20.0 (\textcolor{red}{-12.3}) & 47.1 (\textcolor{red}{-5.3}) & 55.5 (\textcolor{red}{-5.2})
            & 7.9 (\textcolor{red}{-6.4}) & 20.8 (\textcolor{red}{-9.2}) & 27.6 (\textcolor{red}{-9.9}) \\
            \color{red}{\xmark} \color{black}\space IN1k pretrain & 2.3 (\textcolor{red}{-30.0}) & 8.8 (\textcolor{red}{-43.6}) & 14.1 (\textcolor{red}{-46.6})
            & 8.5 (\textcolor{red}{-5.8}) & 22 (\textcolor{red}{-8.0}) & 29.7 (\textcolor{red}{-7.8}) \\
            \color{red}{\xmark} \color{black}\space (K400+IN1k) pretrain & - & - & - & 1.5 (\textcolor{red}{-12.8}) & 4.6 (\textcolor{red}{-25.4}) & 8.1 (\textcolor{red}{-29.4})\\
            \bottomrule
        \end{tabular}
        }
        \caption{Video Swin Transformer ablation study on K400 and  SSv2}
        \label{tab:ablationsK400ssv2}

    \end{center}
\end{table}

\textbf{Removal of Shifted Window:} Instead of computing global attention across all tokens, VST calculates self-attention confined into a smaller local window. While this approach improves computational complexity, it reduces the model’s representation power. This is because the local windows are non-overlapping, and no attention is computed between two windows. To overcome this limitation, VST implements a shifted window approach, which computes self-attention between more tokens belonging to other neighboring windows but is still a much more light-weight operation than global attention. 
In our ablation study, we constrained self-attention to within local windows only and removed the shifted window feature. As expected, the model’s representation power took a hit, as shown in Tab.~\ref{tab:ablationsK400ssv2}.\\
\textbf{Removal of Relative Position Bias:} VST uses relative position bias to represent the distance between the patches withing the local windows. This term is used during the attention computation. Furthermore, Swin \cite{liu2021swin} and VST \cite{liu2021videoswin} skip absolute position embedding. Therefore, removal of the relative position bias robs the model of any position information and therefore causes a stronger performance drop than the removal of the shifted window (Tab.~\ref{tab:ablationsK400ssv2}).

\textbf{No Pretraining:} Removal of ImageNet 1K pretraining from VST has the strongest negative impact on model performance. For the 1\% split, top 1 accuracy drops from 32.3\% to 2.29\%, with the model barely learning anything. A similar behaviour is observed across all splits. These results shows that pretraining is crucial for VTs to perform well on action recognition tasks at low-labeled settings.

\subsection{Importance of Pretraining for VTs versus CNNs}
Next, we investigate the impact of pretraining in CNN performance and compare it to VTs. Tab.~\ref{tab:NopretrainVSTvsCNN} shows the performance comparison for R3D50 and VST without pretraining at 1\% and 10\%. At 1\%, both VST and R3D50 are close; yet at 10\%, R3D50 has substantially higher performance. Ie. at 10\% R3D50 learns much faster than VST and does not suffer as much as VST without pretraining. VST leverages spatial information to video understanding better than CNN or CNN plus Non-local network based methods. 

\begin{table}[h!]
    \begin{center}

        \scalebox{0.9}{
        \begin{tabular}{ccc}
            \toprule
            \textbf{Architecture} & \multicolumn{2}{c}{\textbf{Top-1 Accuracy}} \\
             & 1\% & 10\%\\
            \hline
            R3D-50 in \cite{qian2021cvrl} & 3.2\textsuperscript{\textdagger} & 39.6\textsuperscript{\textdagger} \\
            Slow-R50 \cite{feichtenhofer2019slowfast} & 4.81 & 31.68 \\
            MViTv2 \cite{li2021improvedMViT} Small & 3.1 & 33.68 \\
            Uniformer \cite{li2022uniformer3d} & 4.9 & 27.17 \\
            VST Small \cite{liu2021videoswin} & 2.29 & 14.11 \\
            \bottomrule
        \end{tabular}}
        \caption{CNN vs Transformer performance on K400 without IN1k pretraining. Absent any spatial pretraining CNNs are still more sample efficient on video compared to video transformers.}
        \label{tab:NopretrainVSTvsCNN}
    \end{center}
\end{table}

\subsection{Can VST generalize Uniformer's structure?}
Recall that VST employs windowed local self-attention throughout all of its stages to mitigate computational complexity of attention. This, however, restricts its capacity. This is mitigated by shifting patches to a different local window prior to performing windowed self-attention. This shifting occurs in subsequent self-attention layers in the network. In addition to shifted window approach, patches are merged through the network reducing the number of visual tokens gradually. As the window size of the self-attention is fixed, fewer tokens make subsequent windowed self-attention "more global" creating hierarchical representations deeper in the network. 
 \mbox{}\\ Uniformer, on the other hand, uses 3D convolution for shallow layers to aggregate information from tokens locally and for deeper layers employs typical global self-attention.  The architecture reaches better accuracy using global rather than local attention in these deeper layers. Furthermore, reduced tokens at deeper layers increase efficiency.
 \mbox{}\\
 
\begin{table}[h!]
    \begin{center}
        
        \scalebox{0.9}{
        \begin{tabular}{ccccccccc}
            \toprule
            \textbf{Architecture} & \multicolumn{4}{c}{\textbf{Kinetics Top-1 Accuracy}} & \multicolumn{4}{c}{\textbf{SSv2 Top-1 Accuracy}}\\
             & 1\% & 5\% & 10\% & 100\% & 1\% & 5\% & 10\% & 100\% \\
            \hline
            Uniformer Small & 22.52 & 47.71 & 56.01 & \textbf{80.8} & 12.77 & \textbf{36.14} & \textbf{44.81} & 63.5\\
            VST & 32.3 & 52.37 & 60.66 & 78.39 & \textbf{13.77} & 28.97 & 36.07 & 64.95\\
            VST-LLGG (Ours) & \textbf{32.56} & \textbf{52.78} & \textbf{60.69} & 78.95 & 12.92 & 28.66 & 37.72 & \textbf{65.17} \\
            \bottomrule
        \end{tabular}}
        \caption{VST-LLGG vs VST vs Uniformer performance. Note: Video Swin Small was used for VST and VST-LLGG on K400 and Base for SSv2.}
        \label{tab:LLGG}

    \end{center}
\end{table}

 \looseness=-1  VST's \cite{liu2021videoswin} dominant performance on K400 (Tab.~\ref{tab:K400COMP} compared with Uniformer's \cite{li2022uniformer3d} dominant performance on SSv2 (Tab.~\ref{tab:SSv2COMP}) in low-labeled setting motivated Swin-LLGG. Our observations indicate VST's inductive biases strongly transfer knowledge from spatial pretraining to spatiotemporal classification on datasets that are significantly more spatially dominant such as K400\cite{kay2017kinetics}; this is not the case for SSv2 \cite{goyal2017something} which is less spatially dominant \cite{chen2021deep}. Motivated by this, we investigate whether VST can generalize Uniformer's notion of local attention for early layers followed by global attention for later layers. This was done by relaxing the last 2 stages of VST from local to global attention (VST-LLGG). As shown in Tab. \ref{tab:LLGG} VST-LLGG marginally increases performance on K400 but does not improve performance on SSv2. ie. Swin cannot make up the gap with Uniformer on SSv2 where perhaps 3D convolution as mentioned \hyperref[VTselfcompare]{earlier} may learn finer temporal features that are more relevant for the SSv2 dataset. VST naturally has global attention during the final stage due to patch merging. Hence the representation has already reached a level of hierarchy such that global attention is not as meaningful deeper in the network, ie. local attention on the deeper hierarchical representations of VST is effectively global attention already leading to minor gains with VST-LLGG.

\section{Low-labeled Supervised Training vs Semi-Supervised methods}

\looseness=-1 In low-labeled settings, surprisingly simply using transformers trained on available labeled video data achieved greater performance over semi-supervised methods that use both labeled and unlabeled video data (see Tab.~\ref{Atab:XformerVsSslK00}). Specifically Video Swin transformer by itself achieves greater performance and other transformers are still relatively competitive despite not using any extra unlabeled video data. Typical architectures used in video semi-supervised work tend to be on the lower complexity side (ResNet-18) and an argument can be made this contributes to the performance gap. In addition not all the video SSL methods use CNNs pretrained on IN1K; however they see significantly more video data. Despite these limitations in comparison, the significant gap transformers exhibit is a promising result given the domain shift between images and video. Tab.~\ref{Atab:XformerVsSslK00} 

\begin{table*}[h!]
    \begin{center}
     \scalebox{0.9}{
        \begin{tabular}{ccccccccc}
            \toprule
            \textbf{Method} & \textbf{Backbone} & \multicolumn{4}{c}{\textbf{Top-1 Accuracy}} \\

              & & 1\% & 5\% & 10\% & 100\%\\
            \toprule
            CMPL \cite{xu2022cross} & R3D-50  & 17.6\textsuperscript{\textdagger} & - & 58.4\textsuperscript{\textdagger}  & - \\
            MVPL \cite{xiong2021multiview} & R3D-50   & 17.0\textsuperscript{\textdagger}  & - & 58.2\textsuperscript{\textdagger}  & - \\
            MVPL from \cite{xiao2022temporalgradlearning} & R3D-18   & 9.8\textsuperscript{\textdagger}  & - & 43.8\textsuperscript{\textdagger}  & - \\
            ActorCutMix \cite{zou2021learningActorCutMix} & TSM-R3D-18 & 9.24\textsuperscript{\textdagger} & -& - \\ 
            LfTG \cite{xiao2022temporalgradlearning} & R3D-18 & 9.8\textsuperscript{\textdagger} & - & 43.8\textsuperscript{\textdagger} & - \\
            TCL + ActorCutMix \cite{zou2021learningActorCutMix} & TSM-R3D-18 & 9.02\textsuperscript{\textdagger} & 31.45\textsuperscript{\textdagger} & - & - \\
            TCL \cite{Singh2021} & TSM-R3D-18 \cite{lin2019tsm} & 7.69\textsuperscript{\textdagger} & 30.28\textsuperscript{\textdagger} & -  & - \\
            TCL w/ Fine tune \cite{Singh2021} & TSM-R3D-18 \cite{lin2019tsm} & 8.45\textsuperscript{\textdagger} & 31.5\textsuperscript{\textdagger}  & - & -  \\
            TCL w/ Pretrain \& Finetune \cite{Singh2021} & TSM-R3D-18 \cite{lin2019tsm}   & 11.56\textsuperscript{\textdagger}  & 31.91\textsuperscript{\textdagger} & -  & - \\
            \toprule
            Supervised & I3D-R50  \cite{carreira2017quoi3d}  & 13.73  & 33.49 & 44.37 & 73.5\textsuperscript{\textdaggerdbl} \\
            Supervised & I3D-NL-R50 \cite{wang2018non}  & 13.86 & 34.18 & 44.7 & 74\textsuperscript{\textdaggerdbl} \\
            Supervised & C2D-R50  \cite{wang2018non} & 18.13 & 35.1 & 45.77 & 71.61\textsuperscript{\textdaggerdbl}  \\ 
            Supervised & C2D-NL-R50  \cite{wang2018non} & 19.79  & 38.23  & 46.93 & - \\
            Supervised & R3D-50 from \cite{qian2021cvrl} & 16\textsuperscript{\textdagger} & - & 49.1\textsuperscript{\textdagger}  & 76.0\textsuperscript{\textdagger}  \\
            \hline
            Supervised & ViViT FE \cite{arnab2021vivit} & 24.91 & 41.98 & 48.84 & 78.4\textsuperscript{\textdagger} \\  
            Supervised & ViViT ST \cite{arnab2021vivit} & 23.99 & 40.47 & 46.49 & 79.9\textsuperscript{\textdagger} \\ 
            Supervised & Uniformer Small \cite{li2022uniformer3d} & 22.52 & 47.71 & 56.01 & 80.8\textsuperscript{\textdagger} \\
           Supervised & MViTv2 Small \cite{li2022uniformer3d} & 22.17 & 47.59  & 56.96 & \textbf{81.0}\textsuperscript{\textdagger} \\
            Supervised & Video Swin Small \cite{liu2021videoswin} & \textbf{32.3} & \textbf{52.37} & \textbf{60.66} & 80.6\textsuperscript{\textdagger}  
        \end{tabular}
        }
        \caption{Top 1\% accuracy results on K400 for various low-labeled settings. Supervised indicates backbone was trained on just available labeled data.
        Swin Transformer using no video unlabeled data outperforms CNN based methods using labeled and unlabeled video data. Video Swin especially has significant performance gains on K400. 
        \textsuperscript{\textdagger}Result from respectively cited work. \textsuperscript{\textdaggerdbl}Result from \cite{fan2020pyslowfast}.}
        \label{Atab:XformerVsSslK00}
    \end{center}
\end{table*}

\section{Discussion}
Our results indicate that  Video Transformers (VTs) are better learners in low-labeled video settings than CNNs. While generation 2 video transformers show the greatest sample efficiency, ViViT (natural extension of ViT to video) with greater modeling flexibility also shows greater efficiency on extremely low labeled data splits. This seems counter-intuitive but upon closer inspection is reasonable given information redundancy in video frames. This temporal redundancy leads to spatially pretrained VTs ability to extract reasonable representations for a 3D spatiotemporal token and leverage self-attention to form dynamic feature weighting. Recent video representation learning through self-supervision work has also observed the effects of temporal redundancy, \cite{tong2022videomae, feichtenhofer2022masked} find for masked prediction of autoencoding transformers extremely high masking ratios are optimal when compared with other modalities; 0.9-0.95 for video compared with 0.75 for image \cite{feichtenhofer2022masked} and 0.8 for audio \cite{xu2022masked}. Additionally, as mentioned in Sec. \ref{datasets}, \cite{chen2021deep} note that current large scale video datasets are spatially dominant, SSv2 being less so and K400 being moreso. Our observations alongside these factors point to the now reasonable intuition that spatially training Video Transformers would make them good low-labeled learners. For instance current semi-supervised results in video are 17.6\% Top-1 acc. on 1\% split and 58.4\% for 10\% \cite{xu2022cross}; Swin transformer already surpasses this with no extra unlabeled data, simply fine tuning on available labeled data (Tab. \ref{Atab:XformerVsSslK00}).
CNNs, however catch up in sample efficiency with more data and are still superior when trained tabula rasa (Tab.~\ref{tab:NopretrainVSTvsCNN}). That being said transformers such as MViTv2 and Uniformer are still competitive absent spatial pretraining; a promising result for future work in semi-supervised video recognition.

\section{Conclusion \& Future Work}
Despite the intuition that transformers require substantial data to learn and compensate for CNNs' inductive bias, our extensive experiments show that transformers outperform CNNs in low labeled settings with only a fraction of labeled data. They even significantly outperform, using labeled data alone, more sophisticated semi-supervised methods utilizing CNNs and large unlabeled datasets paired with the labeled data. Through rigorous experimentation, we explain this observation. Specifically, ablation studies on Video Swin Transformer pointed out to substantial impact of pretraining on performance. 

\looseness=-1 Future work includes verifying our findings on multiple other datasets. Furthermore, the analysis of VTs can be expanded to other vision transformer architectures to investigate whether all VTs outperform CNNs in low labeled data settings. With the observation that CNNs updated using learnings from vision transformers \cite{liu2022convnet} also begs the question how updating CNN based architectures would fare in video recognition and specifically low-labeled and semi-supervised settings. Finally, ablation studies performed on Video Swin Transformer could be expanded to other VT architectures to verify that pretraining is crucial for those architectures as well.

\textbf{Acknowledgements}. We thank Yen-Cheng Liu for discussion and comments on our work.

{\small
\bibliography{ssl_video}

\begin{thebibliography}{10}

\bibitem{devlin2018bert}
J.~Devlin, M.-W. Chang, K.~Lee, and K.~Toutanova, ``Bert: Pre-training of deep
  bidirectional transformers for language understanding,'' {\em arXiv preprint
  arXiv:1810.04805}, 2018.

\bibitem{vaswani2017attention}
A.~Vaswani, N.~Shazeer, N.~Parmar, J.~Uszkoreit, L.~Jones, A.~N. Gomez,
  {\L}.~Kaiser, and I.~Polosukhin, ``Attention is all you need,'' {\em Advances
  in neural information processing systems}, vol.~30, 2017.

\bibitem{dosovitskiy2020ViT}
A.~Dosovitskiy, L.~Beyer, A.~Kolesnikov, D.~Weissenborn, X.~Zhai,
  T.~Unterthiner, M.~Dehghani, M.~Minderer, G.~Heigold, S.~Gelly, {\em et~al.},
  ``An image is worth 16x16 words: Transformers for image recognition at
  scale,'' {\em arXiv preprint arXiv:2010.11929}, 2020.

\bibitem{liu2021swin}
Z.~Liu, Y.~Lin, Y.~Cao, H.~Hu, Y.~Wei, Z.~Zhang, S.~Lin, and B.~Guo, ``Swin
  transformer: Hierarchical vision transformer using shifted windows,'' in {\em
  Proceedings of the IEEE/CVF International Conference on Computer Vision},
  pp.~10012--10022, 2021.

\bibitem{carion2020end}
N.~Carion, F.~Massa, G.~Synnaeve, N.~Usunier, A.~Kirillov, and S.~Zagoruyko,
  ``End-to-end object detection with transformers,'' in {\em European
  conference on computer vision}, pp.~213--229, Springer, 2020.

\bibitem{bahdanau2014neural}
D.~Bahdanau, K.~Cho, and Y.~Bengio, ``Neural machine translation by jointly
  learning to align and translate,'' {\em arXiv preprint arXiv:1409.0473},
  2014.

\bibitem{kay2017kinetics}
W.~Kay, J.~Carreira, K.~Simonyan, B.~Zhang, C.~Hillier, S.~Vijayanarasimhan,
  F.~Viola, T.~Green, T.~Back, P.~Natsev, {\em et~al.}, ``The kinetics human
  action video dataset,'' {\em arXiv preprint arXiv:1705.06950}, 2017.

\bibitem{goyal2017something}
R.~Goyal, S.~Ebrahimi~Kahou, V.~Michalski, J.~Materzynska, S.~Westphal, H.~Kim,
  V.~Haenel, I.~Fruend, P.~Yianilos, M.~Mueller-Freitag, {\em et~al.}, ``The"
  something something" video database for learning and evaluating visual common
  sense,'' in {\em Proceedings of the IEEE international conference on computer
  vision}, pp.~5842--5850, 2017.

\bibitem{khan2021transformers}
S.~Khan, M.~Naseer, M.~Hayat, S.~W. Zamir, F.~S. Khan, and M.~Shah,
  ``Transformers in vision: A survey,'' {\em ACM Computing Surveys (CSUR)},
  2021.

\bibitem{liu2021efficient}
Y.~Liu, E.~Sangineto, W.~Bi, N.~Sebe, B.~Lepri, and M.~De~Nadai, ``Efficient
  training of visual transformers with small-size datasets,'' {\em arXiv
  preprint arXiv:2106.03746}, 2021.

\bibitem{el2021large}
A.~El-Nouby, G.~Izacard, H.~Touvron, I.~Laptev, H.~Jegou, and E.~Grave, ``Are
  large-scale datasets necessary for self-supervised pre-training?,'' {\em
  arXiv preprint arXiv:2112.10740}, 2021.

\bibitem{liu2021videoswin}
Z.~Liu, J.~Ning, Y.~Cao, Y.~Wei, Z.~Zhang, S.~Lin, and H.~Hu, ``Video swin
  transformer,'' {\em arXiv preprint arXiv:2106.13230}, 2021.

\bibitem{li2022uniformer3d}
K.~Li, Y.~Wang, P.~Gao, G.~Song, Y.~Liu, H.~Li, and Y.~Qiao, ``Uniformer:
  Unified transformer for efficient spatiotemporal representation learning,''
  {\em arXiv preprint arXiv:2201.04676}, 2022.

\bibitem{arnab2021vivit}
A.~Arnab, M.~Dehghani, G.~Heigold, C.~Sun, M.~Lu{\v{c}}i{\'c}, and C.~Schmid,
  ``Vivit: A video vision transformer,'' in {\em International Conference on
  Computer Vision (ICCV)}, 2021.

\bibitem{li2021improvedMViT}
Y.~Li, C.-Y. Wu, H.~Fan, K.~Mangalam, B.~Xiong, J.~Malik, and C.~Feichtenhofer,
  ``Improved multiscale vision transformers for classification and detection,''
  {\em arXiv preprint arXiv:2112.01526}, 2021.

\bibitem{deng2009imagenet}
J.~Deng, W.~Dong, R.~Socher, L.-J. Li, K.~Li, and L.~Fei-Fei, ``Imagenet: A
  large-scale hierarchical image database,'' in {\em 2009 IEEE conference on
  computer vision and pattern recognition}, pp.~248--255, Ieee, 2009.

\bibitem{sign2021tcl}
A.~Singh, O.~Chakraborty, A.~Varshney, R.~Panda, R.~Feris, K.~Saenko, A.~Das,
  I.~Madras, and I.~Kharagpur, ``Semi-supervised action recognition with
  temporal contrastive learning,''
\newblock arXiv: 2102.02751v2.

\bibitem{xiong2021multiview}
B.~Xiong, H.~Fan, K.~Grauman, and C.~Feichtenhofer, ``Multiview pseudo-labeling
  for semi-supervised learning from video,'' in {\em Proceedings of the
  IEEE/CVF International Conference on Computer Vision}, pp.~7209--7219, 2021.

\bibitem{xiao2022temporalgradlearning}
J.~Xiao, L.~Jing, L.~Zhang, J.~He, Q.~She, Z.~Zhou, A.~Yuille, and Y.~Li,
  ``Learning from temporal gradient for semi-supervised action recognition,''
  in {\em Proceedings of the IEEE/CVF Conference on Computer Vision and Pattern
  Recognition}, pp.~3252--3262, 2022.

\bibitem{terao2022compressedssl}
H.~Terao, W.~Noguchi, H.~Iizuka, and M.~Yamamoto, ``Compressed video ensemble
  based pseudo-labeling for semi-supervised action recognition,'' {\em Machine
  Learning with Applications}, p.~100336, 2022.

\bibitem{jing2021videossl}
L.~Jing, T.~Parag, Z.~Wu, Y.~Tian, and H.~Wang, ``Videossl: Semi-supervised
  learning for video classification,'' in {\em Proceedings of the IEEE/CVF
  Winter Conference on Applications of Computer Vision}, pp.~1110--1119, 2021.

\bibitem{li2022uniformer2d}
K.~Li, Y.~Wang, J.~Zhang, P.~Gao, G.~Song, Y.~Liu, H.~Li, and Y.~Qiao,
  ``Uniformer: Unifying convolution and self-attention for visual
  recognition,'' {\em arXiv preprint arXiv:2201.09450}, 2022.

\bibitem{bertasius2021space}
G.~Bertasius, H.~Wang, and L.~Torresani, ``Is space-time attention all you need
  for video understanding,'' {\em arXiv preprint arXiv:2102.05095}, vol.~2,
  no.~3, p.~4, 2021.

\bibitem{fan2021multiscale}
H.~Fan, B.~Xiong, K.~Mangalam, Y.~Li, Z.~Yan, J.~Malik, and C.~Feichtenhofer,
  ``Multiscale vision transformers,'' in {\em Proceedings of the IEEE/CVF
  International Conference on Computer Vision}, pp.~6824--6835, 2021.

\bibitem{taylor2010convolutional}
G.~W. Taylor, R.~Fergus, Y.~LeCun, and C.~Bregler, ``Convolutional learning of
  spatio-temporal features,'' in {\em European conference on computer vision},
  pp.~140--153, Springer, 2010.

\bibitem{kim2007human}
H.-J. Kim, J.~S. Lee, and H.-S. Yang, ``Human action recognition using a
  modified convolutional neural network,'' in {\em International Symposium on
  Neural Networks}, pp.~715--723, Springer, 2007.

\bibitem{jhuang2007biologically}
H.~Jhuang, T.~Serre, L.~Wolf, and T.~Poggio, ``A biologically inspired system
  for action recognition,'' in {\em 2007 IEEE 11th international conference on
  computer vision}, pp.~1--8, Ieee, 2007.

\bibitem{karpathy2014largeVideo}
A.~Karpathy, G.~Toderici, S.~Shetty, T.~Leung, R.~Sukthankar, and L.~Fei-Fei,
  ``Large-scale video classification with convolutional neural networks,'' in
  {\em Proceedings of the IEEE conference on Computer Vision and Pattern
  Recognition}, pp.~1725--1732, 2014.

\bibitem{ji2012-3d}
S.~Ji, W.~Xu, M.~Yang, and K.~Yu, ``3d convolutional neural networks for human
  action recognition,'' {\em IEEE transactions on pattern analysis and machine
  intelligence}, vol.~35, no.~1, pp.~221--231, 2012.

\bibitem{baccouche2011sequential3dcnnlstm}
M.~Baccouche, F.~Mamalet, C.~Wolf, C.~Garcia, and A.~Baskurt, ``Sequential deep
  learning for human action recognition,'' in {\em International workshop on
  human behavior understanding}, pp.~29--39, Springer, 2011.

\bibitem{tran2015c3d}
D.~Tran, L.~Bourdev, R.~Fergus, L.~Torresani, and M.~Paluri, ``Learning
  spatiotemporal features with 3d convolutional networks,'' in {\em Proceedings
  of the IEEE international conference on computer vision}, pp.~4489--4497,
  2015.

\bibitem{simonyan2014twostream}
K.~Simonyan and A.~Zisserman, ``Two-stream convolutional networks for action
  recognition in videos,'' {\em Advances in neural information processing
  systems}, vol.~27, 2014.

\bibitem{christoph2016spatiotemporal}
R.~Christoph and F.~A. Pinz, ``Spatiotemporal residual networks for video
  action recognition,'' {\em Advances in neural information processing
  systems}, pp.~3468--3476, 2016.

\bibitem{feichtenhofer2017spatiotemporal}
C.~Feichtenhofer, A.~Pinz, and R.~P. Wildes, ``Spatiotemporal multiplier
  networks for video action recognition,'' in {\em Proceedings of the IEEE
  conference on computer vision and pattern recognition}, pp.~4768--4777, 2017.

\bibitem{he2016resnet}
K.~He, X.~Zhang, S.~Ren, and J.~Sun, ``Deep residual learning for image
  recognition,'' in {\em Proceedings of the IEEE conference on computer vision
  and pattern recognition}, pp.~770--778, 2016.

\bibitem{tran2018r3d}
D.~Tran, H.~Wang, L.~Torresani, J.~Ray, Y.~LeCun, and M.~Paluri, ``A closer
  look at spatiotemporal convolutions for action recognition,'' in {\em
  Proceedings of the IEEE conference on Computer Vision and Pattern
  Recognition}, pp.~6450--6459, 2018.

\bibitem{tran2019video}
D.~Tran, H.~Wang, L.~Torresani, and M.~Feiszli, ``Video classification with
  channel-separated convolutional networks,'' in {\em Proceedings of the
  IEEE/CVF International Conference on Computer Vision}, pp.~5552--5561, 2019.

\bibitem{carreira2017quoi3d}
J.~Carreira and A.~Zisserman, ``Quo vadis, action recognition? a new model and
  the kinetics dataset,'' in {\em proceedings of the IEEE Conference on
  Computer Vision and Pattern Recognition}, pp.~6299--6308, 2017.

\bibitem{szegedy2015going}
C.~Szegedy, W.~Liu, Y.~Jia, P.~Sermanet, S.~Reed, D.~Anguelov, D.~Erhan,
  V.~Vanhoucke, and A.~Rabinovich, ``Going deeper with convolutions,'' in {\em
  Proceedings of the IEEE conference on computer vision and pattern
  recognition}, pp.~1--9, 2015.

\bibitem{lin2019tsm}
J.~Lin, C.~Gan, and S.~Han, ``Tsm: Temporal shift module for efficient video
  understanding,'' in {\em Proceedings of the IEEE/CVF International Conference
  on Computer Vision}, pp.~7083--7093, 2019.

\bibitem{feichtenhofer2019slowfast}
C.~Feichtenhofer, H.~Fan, J.~Malik, and K.~He, ``Slowfast networks for video
  recognition,'' in {\em Proceedings of the IEEE/CVF international conference
  on computer vision}, pp.~6202--6211, 2019.

\bibitem{feichtenhofer2020x3d}
C.~Feichtenhofer, ``X3d: Expanding architectures for efficient video
  recognition,'' in {\em Proceedings of the IEEE/CVF Conference on Computer
  Vision and Pattern Recognition}, pp.~203--213, 2020.

\bibitem{yue2015deepcnnlstm}
J.~Yue-Hei~Ng, M.~Hausknecht, S.~Vijayanarasimhan, O.~Vinyals, R.~Monga, and
  G.~Toderici, ``Beyond short snippets: Deep networks for video
  classification,'' in {\em Proceedings of the IEEE conference on computer
  vision and pattern recognition}, pp.~4694--4702, 2015.

\bibitem{donahue2015long}
J.~Donahue, L.~Anne~Hendricks, S.~Guadarrama, M.~Rohrbach, S.~Venugopalan,
  K.~Saenko, and T.~Darrell, ``Long-term recurrent convolutional networks for
  visual recognition and description,'' in {\em Proceedings of the IEEE
  conference on computer vision and pattern recognition}, pp.~2625--2634, 2015.

\bibitem{weston2014memory}
J.~Weston, S.~Chopra, and A.~Bordes, ``Memory networks,'' {\em arXiv preprint
  arXiv:1410.3916}, 2014.

\bibitem{graves2014neural}
A.~Graves, G.~Wayne, and I.~Danihelka, ``Neural turing machines,'' {\em arXiv
  preprint arXiv:1410.5401}, 2014.

\bibitem{wang2018non}
X.~Wang, R.~Girshick, A.~Gupta, and K.~He, ``Non-local neural networks,'' in
  {\em Proceedings of the IEEE conference on computer vision and pattern
  recognition}, pp.~7794--7803, 2018.

\bibitem{guo2022attention}
M.-H. Guo, T.-X. Xu, J.-J. Liu, Z.-N. Liu, P.-T. Jiang, T.-J. Mu, S.-H. Zhang,
  R.~R. Martin, M.-M. Cheng, and S.-M. Hu, ``Attention mechanisms in computer
  vision: A survey,'' {\em Computational Visual Media}, pp.~1--38, 2022.

\bibitem{li2018videolstm}
Z.~Li, K.~Gavrilyuk, E.~Gavves, M.~Jain, and C.~G. Snoek, ``Videolstm
  convolves, attends and flows for action recognition,'' {\em Computer Vision
  and Image Understanding}, vol.~166, pp.~41--50, 2018.

\bibitem{sharma2015action}
S.~Sharma, R.~Kiros, and R.~Salakhutdinov, ``Action recognition using visual
  attention,'' {\em arXiv preprint arXiv:1511.04119}, 2015.

\bibitem{long2018attention}
X.~Long, C.~Gan, G.~De~Melo, J.~Wu, X.~Liu, and S.~Wen, ``Attention clusters:
  Purely attention based local feature integration for video classification,''
  in {\em Proceedings of the IEEE conference on computer vision and pattern
  recognition}, pp.~7834--7843, 2018.

\bibitem{chen20182}
Y.~Chen, Y.~Kalantidis, J.~Li, S.~Yan, and J.~Feng, ``A\^{} 2-nets: Double
  attention networks,'' {\em Advances in neural information processing
  systems}, vol.~31, 2018.

\bibitem{touvron2021training}
H.~Touvron, M.~Cord, M.~Douze, F.~Massa, A.~Sablayrolles, and H.~J{\'e}gou,
  ``Training data-efficient image transformers \& distillation through
  attention,'' in {\em International Conference on Machine Learning},
  pp.~10347--10357, PMLR, 2021.

\bibitem{bulat2021space}
A.~Bulat, J.~M. Perez~Rua, S.~Sudhakaran, B.~Martinez, and G.~Tzimiropoulos,
  ``Space-time mixing attention for video transformer,'' {\em Advances in
  Neural Information Processing Systems}, vol.~34, pp.~19594--19607, 2021.

\bibitem{neimark2021video}
D.~Neimark, O.~Bar, M.~Zohar, and D.~Asselmann, ``Video transformer network,''
  in {\em Proceedings of the IEEE/CVF International Conference on Computer
  Vision}, pp.~3163--3172, 2021.

\bibitem{chen2020big}
T.~Chen, S.~Kornblith, K.~Swersky, M.~Norouzi, and G.~E. Hinton, ``Big
  self-supervised models are strong semi-supervised learners,'' {\em Advances
  in neural information processing systems}, vol.~33, pp.~22243--22255, 2020.

\bibitem{sohn2020fixmatch}
K.~Sohn, D.~Berthelot, N.~Carlini, Z.~Zhang, H.~Zhang, C.~A. Raffel, E.~D.
  Cubuk, A.~Kurakin, and C.-L. Li, ``Fixmatch: Simplifying semi-supervised
  learning with consistency and confidence,'' {\em Advances in neural
  information processing systems}, vol.~33, pp.~596--608, 2020.

\bibitem{kumar2022end}
A.~Kumar and Y.~S. Rawat, ``End-to-end semi-supervised learning for video
  action detection,'' in {\em Proceedings of the IEEE/CVF Conference on
  Computer Vision and Pattern Recognition}, pp.~14700--14710, 2022.

\bibitem{rizve2021defense}
M.~N. Rizve, K.~Duarte, Y.~S. Rawat, and M.~Shah, ``In defense of
  pseudo-labeling: An uncertainty-aware pseudo-label selection framework for
  semi-supervised learning,'' {\em arXiv preprint arXiv:2101.06329}, 2021.

\bibitem{sennrich2015subword}
R.~Sennrich, B.~Haddow, and A.~Birch, ``Neural machine translation of rare
  words with subword units,'' {\em arXiv preprint arXiv:1508.07909}, 2015.

\bibitem{gage1994BPE}
P.~Gage, ``A new algorithm for data compression,'' {\em C Users Journal},
  vol.~12, no.~2, pp.~23--38, 1994.

\bibitem{carreira2017i3d}
J.~Carreira and A.~Zisserman, ``Quo vadis, action recognition? a new model and
  the kinetics dataset,'' in {\em proceedings of the IEEE Conference on
  Computer Vision and Pattern Recognition}, pp.~6299--6308, 2017.

\bibitem{chen2021deep}
C.-F.~R. Chen, R.~Panda, K.~Ramakrishnan, R.~Feris, J.~Cohn, A.~Oliva, and
  Q.~Fan, ``Deep analysis of cnn-based spatio-temporal representations for
  action recognition,'' in {\em Proceedings of the IEEE/CVF Conference on
  Computer Vision and Pattern Recognition}, pp.~6165--6175, 2021.

\bibitem{qian2021cvrl}
R.~Qian, T.~Meng, B.~Gong, M.-H. Yang, H.~Wang, S.~Belongie, and Y.~Cui,
  ``Spatiotemporal contrastive video representation learning,'' in {\em
  Proceedings of the IEEE/CVF Conference on Computer Vision and Pattern
  Recognition}, pp.~6964--6974, 2021.

\bibitem{fan2020pyslowfast}
H.~Fan, Y.~Li, B.~Xiong, W.-Y. Lo, and C.~Feichtenhofer, ``Pyslowfast.''
  \url{https://github.com/facebookresearch/slowfast}, 2020.

\bibitem{park2022howDoVTsWork}
N.~Park and S.~Kim, ``How do vision transformers work?,'' {\em arXiv preprint
  arXiv:2202.06709}, 2022.

\bibitem{xu2022cross}
Y.~Xu, F.~Wei, X.~Sun, C.~Yang, Y.~Shen, B.~Dai, B.~Zhou, and S.~Lin,
  ``Cross-model pseudo-labeling for semi-supervised action recognition,'' in
  {\em Proceedings of the IEEE/CVF Conference on Computer Vision and Pattern
  Recognition}, pp.~2959--2968, 2022.

\bibitem{zou2021learningActorCutMix}
Y.~Zou, J.~Choi, Q.~Wang, and J.-B. Huang, ``Learning representational
  invariances for data-efficient action recognition,'' {\em arXiv preprint
  arXiv:2103.16565}, 2021.

\bibitem{Singh2021}
A.~Singh, O.~Chakraborty, A.~Varshney, R.~Panda, R.~Feris, K.~Saenko, A.~Das,
  I.~Madras, and I.~Kharagpur, ``Semi-supervised action recognition with
  temporal contrastive learning,'' 2021.

\bibitem{tong2022videomae}
Z.~Tong, Y.~Song, J.~Wang, and L.~Wang, ``Videomae: Masked autoencoders are
  data-efficient learners for self-supervised video pre-training,'' {\em arXiv
  preprint arXiv:2203.12602}, 2022.

\bibitem{feichtenhofer2022masked}
C.~Feichtenhofer, H.~Fan, Y.~Li, and K.~He, ``Masked autoencoders as
  spatiotemporal learners,'' {\em arXiv preprint arXiv:2205.09113}, 2022.

\bibitem{xu2022masked}
H.~Xu, J.~Li, A.~Baevski, M.~Auli, W.~Galuba, F.~Metze, C.~Feichtenhofer, {\em
  et~al.}, ``Masked autoencoders that listen,'' {\em arXiv preprint
  arXiv:2207.06405}, 2022.

\bibitem{liu2022convnet}
Z.~Liu, H.~Mao, C.-Y. Wu, C.~Feichtenhofer, T.~Darrell, and S.~Xie, ``A convnet
  for the 2020s,'' in {\em Proceedings of the IEEE/CVF Conference on Computer
  Vision and Pattern Recognition}, pp.~11976--11986, 2022.

\bibitem{2020mmaction2}
M.~Contributors, ``Openmmlab's next generation video understanding toolbox and
  benchmark.'' \url{https://github.com/open-mmlab/mmaction2}, 2020.

\bibitem{dehghani2021scenic}
M.~Dehghani, A.~Gritsenko, A.~Arnab, M.~Minderer, and Y.~Tay, ``Scenic: A jax
  library for computer vision research and beyond,'' in {\em Proceedings of the
  IEEE/CVF Conference on Computer Vision and Pattern Recognition (CVPR)},
  pp.~21393--21398, 2022.

\end{thebibliography}
}
\clearpage
\makeatother

\appendix
\section{Appendix}

\setcounter{table}{0}
\renewcommand{\thetable}{A\arabic{table}}

\begin{table*}[h!]
    
    \begin{center}
    \resizebox{\linewidth}{!}{
        \begin{tabular}{ccccccccc}
            \toprule
            \textbf{Architecture} & \multicolumn{4}{c}{\textbf{Top-1 Accuracy}} & \multicolumn{4}{c}{\textbf{Top-5 Accuracy}} \\

             & 1\% & 5\% & 10\% & 100\% & 1\% & 5\% & 10\% & 100\%\\
            \toprule
            I3D-R50  \cite{carreira2017quoi3d}  & 13.73(\textcolor{red}{-3.79})  & 33.49 & 44.37(\textcolor{red}{-4.73})  & 73.5\textsuperscript{\textdaggerdbl} (\textcolor{red}{-2.5}) & 29.39 & 57 & 68.59 & 90.8\textsuperscript{\textdaggerdbl}  \\
            I3D-NL-R50 \cite{wang2018non}  & 13.86 (\textcolor{red}{-2.14}) & 34.18 (\textcolor{red}{-0.69}) & 44.7 (\textcolor{red}{-4.4}) & 74\textsuperscript{\textdaggerdbl} (\textcolor{red}{-2}) & 30.29 & 57.48 & 68.95 & 91.1\textsuperscript{\textdaggerdbl} \\
            C2D-R50  \cite{wang2018non} & 18.13 (\textcolor{red}{-1.66}) & 35.1 (\textcolor{red}{-3.13}) & 45.77 (\textcolor{red}{-3.33})  & 71.61\textsuperscript{\textdaggerdbl} (\textcolor{red}{-4.39}) & 36.34 & 58.91 & 69.39 & 87.8\textsuperscript{\textdaggerdbl}\\
            C2D-NL-R50  \cite{wang2018non} & \textbf{19.79}  & \textbf{38.23}  & 46.93 ( (\textcolor{red}{-2.71}) & - & 39.35 & 61.96 & 70.88 & -\\
            R3D-50 from \cite{qian2021cvrl} & 16\textsuperscript{\textdagger} (\textcolor{red}{-2.13}) & - & \textbf{49.1}\textsuperscript{\textdagger}  & \textbf{76.0}\textsuperscript{\textdagger}  & - & - & - & - \\
            SLOWFAST-R3D-50 \cite{feichtenhofer2019slowfast} & 6.09 (\textcolor{red}{-13.7}) & - & 36.07(\textcolor{red}{-13.03}) & - & 15.86 & - & 59.7 & - \\
            \hline
            ViViT FE \cite{arnab2021vivit} & 24.91 (\textcolor{blue}{+5.12}) & 41.98 (\textcolor{blue}{+3.75})& 48.84 (\textcolor{red}{-0.26}) & 78.4\textsuperscript{\textdagger} (\textcolor{blue}{+2.4}) & 46.43 & 65.55 & 72.15 & -\\
            ViViT ST \cite{arnab2021vivit} & 23.99 (\textcolor{blue}{+4.2}) & 40.47 (\textcolor{blue}{2.24}) & 46.49 (\textcolor{red}{-2.61}) & 79.9\textsuperscript{\textdagger} (\textcolor{blue}{+3.9}) & 46.25 & 64.45 & 70.56 & - \\
            Uniformer Small \cite{li2022uniformer3d} & 22.52 (\textcolor{blue}{+2.73}) & 47.71 (\textcolor{blue}{+9.48}) & 56.01 (\textcolor{blue}{+6.91}) & 80.8\textsuperscript{\textdagger} (\textcolor{blue}{+4.8}) & 43.59 & 71.06 & 77.99 & 94.7\textsuperscript{\textdagger}\\
           MViTv2 Small \cite{li2022uniformer3d} & 22.17 (\textcolor{blue}{+2.38}) & 47.59  (\textcolor{blue}{+9.36}) & 56.96 (\textcolor{blue}{+7.86}) & 81.0 \textsuperscript{\textdagger} (\textcolor{blue}{+5}) & 43.17 & 70.94 & 79.57 & 94.6\textsuperscript{\textdagger}\\
            Video Swin Small \cite{liu2021videoswin} & 32.3 (\textcolor{blue}{+12.56}) & 52.37 (\textcolor{blue}{+14.14}) & 60.66 (\textcolor{blue}{+11.56}) & 80.6\textsuperscript{\textdagger} (\textcolor{blue}{+4.6}) & 56.09 & 75.64 & 80.06 & 94.5\textsuperscript{\textdagger}\\        \bottomrule
        \end{tabular}
        }
        \caption{Top 1\% \& Top 5\% accuracy results on K400 for various low-labeled settings for CNN-based (top half) vs Transformer-based (bottom half). 
        Differences (for top-1) are computed against the highest performing CNN-based model for each split in \textbf{bold}. 
        \textsuperscript{\textdagger}Result from respectively cited work. \textsuperscript{\textdaggerdbl}Result from \cite{fan2020pyslowfast}.}
         \label{Atab:K400COMP}
    \end{center}
\end{table*}
\begin{table*}[h!]

    \begin{center}
        
        \resizebox{\linewidth}{!}{%
        \begin{tabular}{ccccccccc}
            \toprule
            \textbf{Architecture} & \multicolumn{4}{c}{\textbf{Top-1 Accuracy}} & \multicolumn{4}{c}{\textbf{Top-5 Accuracy}} \\
             & 1\% & 5\% & 10\% & 100\% & 1\% & 5\% & 10\% & 100\% \\
            \hline
            I3D-R50  \cite{carreira2017quoi3d} & 8.47 (\textcolor{red}{-1.59})  & 24.3 (\textcolor{red}{-2.5}) & 32.92(\textcolor{red}{-3.25})  & - & 24.19 & 51.52 & 61.81 & -\\
            I3D-NL-R50  \cite{wang2018non} & 8.42(\textcolor{red}{-1.64})  & 24.85(\textcolor{red}{-1.95}) & 32.59(\textcolor{red}{-3.58}) & - & 23.22 & 50.92 & 62.11 & -\\
            C2D-R50  \cite{wang2018non} & 7.16 (\textcolor{red}{-2.9})  & - & 20.16 (\textcolor{red}{-16.01}) & - & 20.16 & - & 47.05 & - \\
            C2D-NL-R50  \cite{wang2018non} & 7.31 (\textcolor{red}{-2.75})  & - & 21.19 (\textcolor{red}{-16.04}) & - & 20.62 & - & 49.3 & - \\
            SlowFast-R50  \cite{feichtenhofer2019slowfast} & \textbf{10.06}  & \textbf{26.8} & \textbf{36.17} & \textbf{63}\textsuperscript{\textdagger} & 25.49 & 53.88 & 64.18 & 88.5\textsuperscript{\textdaggerdbl}\\
            \hline
            Video Swin Base \cite{liu2021videoswin} & 14.3 (\textcolor{blue}{+4.11}) & 30.0 (\textcolor{blue}{+3.2}) & 37.5(\textcolor{blue}{+1.33}) & 65.67 (\textcolor{blue}{+2.67}) & 36.24 & 62.52 & 71.22 & 90.65\\
            Uniformer Small \cite{li2022uniformer3d} & 12.77 (\textcolor{blue}{+2.9})  & 36.14 (\textcolor{blue}{+9.34}) & 44.81 (\textcolor{blue}{+8.64}) & 63.5\textsuperscript{\textdagger}(\textcolor{blue}{+0.5}) & 31.97 & 65.36 & 74.5 & 85.75\\
            \bottomrule
        \end{tabular}}
        \caption{Top 1\% accuracy on SSv2 for various low-labeled settings. CNN-based (top half) vs. Transformer-based (bottom half). Differences (for top-1) are computed against baseline  CNN-based model in \textbf{bold}. \textsuperscript{\textdagger}Result obtained from respectively cited work. \textsuperscript{\textdaggerdbl}Result obtained from \cite{fan2020pyslowfast}.} 
        \label{Atab:SSv2COMP}

    \end{center}
\end{table*}

\subsection{CNN Selection}
\looseness=-1 We selected CNNs across a variety of inductive biases (spatial vs spatiotemporal) that show relatively comparable performance at 100\% but also would be reasonable in low-labeled settings. The architectures we selected range both 2D and 3D CNNs, some of which are also common architectures used in current semi-supervised video recognition work (Tab. \ref{Atab:XformerVsSslK00}). The motivation to show that in the \textit{reasonable best case} CNN based architectures are outperformed by Vision Transformers in low-labeled settings. For instance SlowFast \cite{feichtenhofer2019slowfast}is a good low-labeled candidate for SSv2 (Tab.~\ref{Atab:SSv2COMP} but not as much for K400 (Tab.~\ref{Atab:K400COMP}). When SlowFast is boostrapped from ResNet \cite{he2016resnet}, the fusion layers between the slow and fast path are still trained from scratch. As mentioned by \cite{feichtenhofer2019slowfast} SlowFast generalizes when trained from scratch vs initialization from large scale spatial pretraining (IN1K), however in the low-labeled setting this additional pretraining matters. As such, SlowFast once trained on K400 is in better parameter space (fusion params \& both pathways) making it a better candidate for downstream transfer for Low-labeled SSv2 settings. This was verified through experiments as shown in Tab. \ref{Atab:K400COMP} \& \ref{Atab:SSv2COMP}. Similarly the "type" of video dataset is also a factor, C2D with NL layers \cite{wang2018non}  is the best performing CNN on K400 low-labeled because the spatial pretraining combined with an attention-like mechanism is a strong prior for spatially dominant datasets such as K400. While SSv2 is still spatially dominant \cite{chen2021deep}, it is less so compared with K400. Greater temporal modeling is required hence SlowFast and I3D outperforming the 2D counterparts. In addition the 2D architectures start off efficiently however do not have the capacity to make use of the full dataset and do not perform as well with full data.

\section{Implementation/Training details}
\looseness=-1 For experiments we used the Video Swin official implementation from \cite{liu2021videoswin} through MMaction2 \cite{2020mmaction2}, Uniformer official implementation from \cite{li2022uniformer3d} through \cite{fan2020pyslowfast}, ViViT \cite{arnab2021vivit} official implementation from Scenic \cite{dehghani2021scenic}, MViTv2 from implementation in \cite{fan2020pyslowfast}. CNN experiments were also using implementations in \cite{fan2020pyslowfast}. For training and inference we used hyperparameters and settings as described in each work \cite{li2022uniformer3d, liu2021videoswin, li2021improvedMViT, arnab2021vivit}.
\end{document}